\ifdefined\XeTeXrevision\else\pdfoutput=1\fi
\documentclass{article}
\usepackage{iclr2027_conference,times}

\usepackage{amsmath,amsfonts,bm}

\def\eqref#1{equation~\ref{#1}}
\def\1{\bm{1}}

\def\vx{{\bm{x}}}

\def\mX{{\bm{X}}}

\DeclareMathAlphabet{\mathsfit}{\encodingdefault}{\sfdefault}{m}{sl}
\SetMathAlphabet{\mathsfit}{bold}{\encodingdefault}{\sfdefault}{bx}{n}

\newcommand{\R}{\mathbb{R}}

\usepackage{hyperref}
\usepackage{url}
\usepackage{graphicx}
\usepackage{booktabs}
\usepackage{amsmath,amssymb}
\usepackage{xcolor}
\usepackage[capitalise,noabbrev]{cleveref}
\usepackage{microtype}
\usepackage{float}

\newcommand{\fpt}{\texttt{fp32}}
\newcommand{\bft}{\texttt{bf16}}
\newcommand{\fps}{\texttt{fp16}}
\newcommand{\pp}{\,pp}

\title{Same Probe, Different Numbers: Are Activation Probes Robust to Inference-Time Numerical Non-Determinism?}

\author{Alizishaan Khatri \\
Wrynx Inc. \\
\texttt{research@wrynx.com}}

\iclrfinalcopy

\begin{document}

\maketitle

\begin{abstract}
Activation probes are increasingly used to monitor LLMs in deployment.
A probe is typically trained once per model and dataset under one inference configuration, then used under whatever batch size and numerical precision the serving stack happens to use.
Because common GPU kernels are not batch-invariant and floating-point formats round differently, the activations seen at deployment are not the ones the probe was trained on.
We measure what that costs, for Llama-3.1-8B, Qwen3-8B and Gemma-3-4B, across batch sizes 4, 8 and 16 and float32, bfloat16 and float16, at four depths and four token positions.
We train 768 probes on one configuration, evaluate each on every other, and compare the same probe's verdicts between two runs example by example.

Our central finding is that probes are stable under these perturbations, but that aggregate accuracy is the wrong instrument for showing it: it understates how many individual verdicts change by a factor of two to nine, and that churn, once measured, proves symmetric and leaves the probe's ranking intact.
At the prompt, probe accuracy never moves by more than 0.47 percentage points across 1{,}392 transfers, and only 0.076\% of individual verdicts change; under float32 with only the batch size varied, not one verdict changes in 201{,}960 evaluations.
During decoding the flip rate rises to 2.8\%, but it separates cleanly: rows whose realised tokens matched flip in 0.12--0.15\% of cases at every position, while rows whose tokens diverged flip in 12.9\%.
Reconstructing each run's greedy trajectory from stored logits shows why, and shows that the cause is the text rather than the arithmetic: a bfloat16 batch-size change flips the first generated token for 2.1\% of rows and leaves the two runs on different tokens in 25\% of rows by token 20, while float32 diverges nowhere.
Flips are symmetric between classes, Cohen's $\kappa$ stays above 0.94, and AUROC moves by at most 0.05 points, so what the perturbation disturbs is the handful of examples within rounding distance of the boundary.

Underneath, the activations move about as much as the format's rounding: in bfloat16 a batch-size change perturbs them by a median relative $\ell_2$ of $10^{-2}$, roughly $8\times$ the same change in float16, matching float16's three extra mantissa bits, and for Gemma, where both runs share a GPU, more than the entire float32-to-bfloat16 conversion.
Probes absorb this; the model's own next-token argmax does not.
Robustness evaluations of activation monitors should therefore report per-example agreement rather than aggregate accuracy, separate representational noise from input change, and state the serving configuration alongside the result.
We will release our extraction and analysis pipeline.

\end{abstract}

% ---------------------------------------------------------------------------
\section{Introduction}
\label{sec:intro}

Probing classifiers trained on a language model's hidden states have moved from an analysis tool \citep{alain2016probes,tenney2019bert,belinkov2022probing} to a deployment tool.
Linear and shallow nonlinear probes read out truthfulness \citep{azaria2023lying,marks2024geometry,orgad2025llmsknow}, detect deception and sleeper-agent defection \citep{goldowskydill2025deception,macdiarmid2024probes}, flag high-stakes interactions \citep{mckenzie2025highstakes}, gate autonomous systems on latent evidence of harm \citep{khatri2026safety}, and now run as production misuse monitors \citep{kramar2026probes,cunningham2025costeffective}.
Their appeal is cost: a probe reuses activations the model computes anyway.

Deployment creates a mismatch that has received little attention.
A probe is trained once per model, task and precision, from activations extracted under one inference configuration.
At serving time the same model runs with whatever batch size the scheduler assembles and often at a different numerical precision.
Neither change is semantically meaningful, yet both change the numbers.
Many GPU kernels (matrix multiplication, normalization, attention) are not batch-invariant: their reduction order, and therefore their rounding, depends on the batch shape \citep{he2025nondeterminism,pytorch_reproducibility}.
Lower-precision formats round differently again \citep{micikevicius2018mixed,kalamkar2019bfloat16}.
Recent work shows that these effects are large enough to change greedy LLM outputs and benchmark scores \citep{yuan2025nondeterminism,atil2024nondeterminism}.
Whether they also break the probes that read those models' internals has not been measured.

We run a controlled study.
With a fixed software stack, and on one GPU except where a model's \fpt{} weights did not fit (\cref{sec:setup}), we extract residual-stream activations from Llama-3.1-8B \citep{grattafiori2024llama3}, Qwen3-8B \citep{yang2025qwen3} and Gemma-3-4B \citep{gemmateam2025gemma3}.
We cover three datasets, three batch sizes, three precisions, four depths, and four token positions: the last prompt token plus three steps of greedy decoding.
We characterise how activations differ across configurations, then train 768 probes of three architectures and evaluate each across configurations.
Our findings are:

\begin{itemize}
\item \textbf{At the prompt, numerical noise is small, has no systematic direction, and is harmless to probes.}
Under \fpt, prefill activations of Llama and Qwen are bitwise identical across batch sizes.
Under \bft, changing only the batch size perturbs activations about $8\times$ more than the same change under \fps, matching their mantissa widths, and (for Gemma, where the GPU is held fixed) at least as much as the \fpt$\to$\bft{} conversion itself, while preserving global geometry (CKA $\geq 0.995$).
Across 1{,}392 prompt-position transfers, probe accuracy never moves by more than 0.47\pp{} and shows no directional bias (\cref{sec:acts-prefill,sec:probes}).
\item \textbf{During decoding, rounding noise becomes different text.}
Comparing realised tokens directly, a \bft{} batch-size change flips the first generated token for 2.1\% of rows and leaves the two runs on different tokens in 25\% of rows by token 20; under \fpt{} not one of 5{,}610 rows per model diverges anywhere (\cref{sec:acts-decode}).
Probe accuracy at decode positions changes in 74\% of transfers, by up to 2.35\pp, growing with the diverged fraction (Spearman $\rho=0.73$; \cref{sec:probes}).
\item \textbf{The probe is robust; its input is not.}
Measuring per-example agreement rather than accuracy, 0.076\% of verdicts change at the prompt, where both runs read identical text, against 12.9\% on rows whose generated tokens diverged, and exactly zero in 201{,}960 \fpt{} evaluations.
Rows whose tokens matched flip in 0.12--0.15\% of cases at every decode position, so the instability follows the text, not the arithmetic (\cref{sec:probes}).
\end{itemize}

\noindent One distinction organises all of this: a numerical perturbation can be \emph{benign}, changing the representation but no decision; \emph{behaviourally consequential}, changing a discrete choice and hence the model's future input; or a \emph{validity failure}, where the format cannot represent the activations at all.
Our prompt, decode and excluded-\fps{} cases are one of each.
Our title asks whether probes are robust to inference-time numerical non-determinism, and the answer has two parts.
For a fixed input, yes: probes on prompt representations can be trained under one batch size and precision and deployed under another.
Under autoregressive generation, the probe remains robust but the input does not: monitors that read activations of generated tokens should expect their input text, and therefore their verdicts, to vary with serving configuration unless decoding is made batch-invariant.
Robustness evaluations of activation monitors should separate these two effects, which is the methodological point we want to make.

% ---------------------------------------------------------------------------
\section{Background and related work}
\label{sec:related}

\paragraph{Numerical non-determinism in inference.}
Floating-point addition is not associative, so a kernel that splits a reduction differently for different batch shapes returns slightly different results for the same row.
\citet{he2025nondeterminism} identify this lack of batch invariance in normalization, matrix multiplication and attention as the main source of temperature-0 non-determinism in LLM serving and provide batch-invariant kernels, which serving frameworks have begun to adopt \citep{lmsys2025deterministic}.
\citet{yuan2025nondeterminism} show that batch size, GPU count and type, and \bft{} versus \fpt{} change reasoning-model outputs and accuracy under greedy decoding.
\citet{atil2024nondeterminism} and \citet{song2024greedy} document run-to-run variation in nominally deterministic LLM evaluation, and in training, implementation-level non-determinism alone produces variance comparable to changing the random seed \citep{pham2020variance,zhuang2022randomness,summers2021nondeterminism}.
All of this work studies outputs or trained weights; we study the intermediate representations that downstream probes consume.

\paragraph{Precision formats and outlier activations.}
\fps{} has 5 exponent and 10 mantissa bits (maximum 65{,}504); \bft{} trades precision for \fpt's dynamic range with 8 exponent and 7 mantissa bits \citep{kalamkar2019bfloat16}.
LLM residual streams contain a few very large activations \citep{dettmers2022llmint8,sun2024massive,kovaleva2021bertbusters,puccetti2022outliers,bondarenko2023quantizable}, which can exceed the \fps{} range as happened with T5 \citep{raffel2020t5,hf2020t5fp16overflow}.
Work on quantization and interpretability \citep{singh2025quantization} examines how low-bit weights alter internal structure; we instead keep the weights fixed and vary only the compute format and batch shape.

\paragraph{Probes and their robustness.}
Probes are evaluated mainly for selectivity \citep{hewitt2019control}, for what they reveal about representations \citep{belinkov2022probing,park2024linear,gurnee2024spacetime,gurnee2023haystack}, and for generalization across datasets and distributions \citep{marks2024geometry,orgad2025llmsknow,kramar2026probes}; similarity measures such as CKA \citep{kornblith2019cka} and SVCCA \citep{raghu2017svcca} compare representations across models or runs.
Closest to our setting, \citet{khatri2026reproducibility} reproduce latent-space safety probes across model families and report prompt activations to be largely unperturbed by batch size, which our prompt-position results corroborate and extend to decoding.
To our knowledge, no prior work isolates the shift induced purely by inference-time numerics and measures its effect on probe behaviour.

% ---------------------------------------------------------------------------
\section{Experimental setup}
\label{sec:setup}

\paragraph{Models and data.}
We use Llama-3.1-8B (base), Qwen3-8B (post-trained) and Gemma-3-4B (instruction-tuned) from the Hugging Face Hub \citep{wolf2020transformers}.
Datasets are Rotten Tomatoes sentiment (10{,}662 sentences; the official 8{,}530/1{,}066/1{,}066 split; \citealp{pang2005seeing}), a de-duplicated SMS Spam Collection (5{,}159 messages; \citealp{almeida2011sms}) and the 817 questions of TruthfulQA \citep{lin2022truthfulqa}.
SMS Spam and TruthfulQA are split 70/15/15 by a hash of each row's content, so every run sees identical splits.
Every dataset is pinned to a fixed repository revision.
TruthfulQA has no binary label under our preprocessing, so we use it for activation-level analysis only.
Raw text is fed without a chat template, left-padded to 128 tokens.

\paragraph{Configuration grid.}
For every model and dataset we run the full grid of batch sizes $B\in\{4,8,16\}$ and precisions $\{\fpt,\bft,\fps\}$, with weights loaded directly in the target dtype.
All runs use one seed (42), the same row order, deterministic PyTorch algorithms \citep{paszke2019pytorch} with a fixed cuBLAS workspace, and a single fixed software environment.
Deterministic mode suppresses run-to-run noise at a fixed shape; it does not make kernels batch-invariant.

\paragraph{Extraction hardware, and what it confounds.}
Most runs used one NVIDIA L4 GPU on Google Colab.
The exception is the two 8B models at \fpt{}: their \fpt{} weights exceed the L4's memory, so those runs used an A100.
Every Gemma-3-4B run, and every \bft{} and \fps{} run of the 8B models, used the L4.
This makes two classes of comparison.
\emph{Same-GPU} comparisons change only batch size or precision: all batch-size comparisons, all Gemma comparisons, and all \bft{}$\leftrightarrow$\fps{} probe transfers.
\emph{Cross-GPU} comparisons, namely \fpt{} versus \bft{} or \fps{} for Llama and Qwen, change GPU and precision together, so they bound the combined effect; we never rest a precision-specific claim on one.
Both classes are labelled in every table, and \cref{tab:gpu-split} (\cref{app:gpu}) repeats the probe results split by class.
After the prompt forward pass, we decode 20 tokens greedily with a KV cache.
We record the hidden state after block 3 and after the blocks at 50\%, 75\% and 100\% of depth (Llama 16/24/32, Qwen 18/27/36, Gemma 17/26/34).
We record these at four positions: the last prompt token (\emph{prefill}) and generated tokens 5, 10 and 20 (\emph{decode}).
All vectors are upcast to \fpt{} for storage.

\paragraph{Excluded configuration: Gemma-3-4B in \fps.}
\label{sec:gemma}
Gemma-3-4B run in \fps{} reproduces a known numerical instability.
Its residual activations exceed the \fps{} range and become non-finite.
Community reports trace this to the embedding scaling combined with large post-normalization weights \citep{hf2025gemma3fp16issue,hf2025gemma3fp16clip,hf2025gemma3activationscaling,unsloth2025gemma3}.
In our runs, every \fps{} hidden state at blocks 17, 26 and 34 and every logit vector is NaN, at all four positions, on every dataset and batch size (80\% of stored tensors).
Greedy decoding from NaN logits also makes the finite block-3 decode activations meaningless.
We therefore discard every activation comparison, geometry metric and probe result that involves Gemma \fps{}, whether as the training or the evaluation configuration.
All numbers in this paper cover Llama and Qwen at all three precisions and Gemma at \fpt{} and \bft{}.

\paragraph{Activation-level metrics.}
For two runs $a,b$ differing in one factor we compare the same row's vectors $\vx_a,\vx_b\in\R^{d}$ at the same layer and position, using relative $\ell_2$ distance $\|\vx_a-\vx_b\|_2/\|\vx_a\|_2$, cosine similarity, the fraction of bitwise-identical rows, and the median element-wise ULP distance (how many representable \fpt{} values separate two elements).
At the population level we use linear CKA \citep{kornblith2019cka}, the orthogonal-Procrustes residual and 10-nearest-neighbour Jaccard overlap (\cref{app:metrics}).
Row-level metrics use up to 3{,}000 rows per model and dataset; kNN overlap uses 1{,}500.
Activations at decode positions depend on the generated prefix.
Token ids were not logged during extraction, so the activation-level metrics below use a distance criterion: we call a row \emph{diverged at a position} if its relative $\ell_2$ at block 3 exceeds 0.1 there.
The next paragraph recovers the realised tokens themselves from the stored logits, and \cref{sec:acts-decode} validates the criterion against them.
Block 3 is dominated by the identity of the token currently being processed, and its per-row distances are clearly bimodal: numerical noise sits near $10^{-2.5}$ and diverged rows near $10^{0}$ (\cref{app:fork}).
The criterion detects that two runs process different tokens at that step, not the step at which they first parted.

\paragraph{Measuring token-level divergence.}
\label{sec:tokensweep}
Activations at decode positions depend on the generated prefix, so we also measure that prefix directly.
Extraction stored the output logits at every logged position, and decoding is greedy, so each run's realised token at a position is the argmax of its stored logit vector; no model needs to be run again.
For every pair of runs of the same model and dataset we compare these tokens row by row on the test split, giving per-position token agreement and, for each row, the first logged position at which the two runs disagree.
This covers 174 run pairs after the \fps{} Gemma exclusion.
Since only positions 5, 10 and 20 are logged, a divergence that appears and self-corrects in between is invisible, and we locate the first \emph{logged} disagreement rather than the true first one.
The same sweep also evaluates the existing probe checkpoints at the final layer and at each of the four logged positions in turn, in two ways: on the two row subsets it defines (trajectories agreeing everywhere versus not), and by running each probe over \emph{both} runs' activations for the same rows, which gives the per-example rate at which its verdict changes.

\paragraph{Probes and transfer protocol.}
We train three probe families: logistic regression (LR), an MLP with one hidden layer of 256 units (MLP-1), and an MLP with three hidden layers of 256 units (MLP-3).
The MLPs use ReLU and dropout 0.1. All probes are trained with the cross-entropy loss and Adam (learning rate $10^{-3}$, minibatch 256), up to 100 epochs, and early stopping on validation loss with patience 10.
Inputs are standardized with training-set statistics.
Every probe is trained on the train split at $B=4$ and one precision, for one model, dataset, layer and position, giving $3\times2\times3\times4\times4\times3=864$ probes, of which 768 remain after the exclusion above.
Each probe keeps its standardization and is evaluated on the test split of every available (precision, batch size) cell: nine per probe for Llama and Qwen (5{,}184 evaluations) and six for Gemma (1{,}152).
We report $\Delta = \mathrm{acc}(\text{test cell}) - \mathrm{acc}(\text{matched cell})$, where the matched cell has the probe's own training configuration.
This yields 5{,}568 transfer measurements.
Because the evaluation examples are identical across cells, $|\Delta|\cdot n$ lower-bounds the number of flipped predictions.
As a control on the probe pipeline itself, we re-ran the entire sweep from scratch on the same stored activations.
The re-run covered the unfiltered grid of 864 probes $\times$ 9 cells $=7{,}776$ (probe, cell) rows, of which 6{,}336 survive the \fps{} Gemma exclusion, and reproduced both the accuracy and the macro-F1 of every row exactly.
Probe training is therefore deterministic here at a fixed seed, and every difference we report comes from the extraction configuration.
This control says nothing about how much a probe would move under a \emph{different} training seed, which we did not measure (\cref{sec:discussion}).
Matched accuracies (\cref{app:matched}) range from 74--87\% on Rotten Tomatoes and 91--99\% on SMS Spam across probe types and positions.
They differ by at most 0.4\pp{} between training precisions, so no precision is a clearly worse training choice.

% ---------------------------------------------------------------------------
\section{How much do activations change?}
\label{sec:acts}

\begin{figure}[t]
\centering
\includegraphics[width=\linewidth]{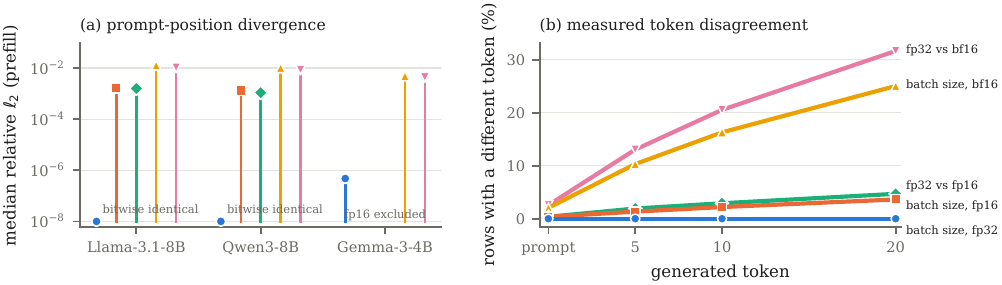}
\caption{\textbf{(a)} Median per-row relative $\ell_2$ distance at the last prompt token, pooled over layers and datasets.
Stems start at the $10^{-8}$ axis floor; Llama and Qwen \fpt{} runs are bitwise identical across batch sizes, and Gemma-3-4B has no \fps{} comparisons (\cref{sec:gemma}).
\textbf{(b)} Percentage of test rows whose realised token differs between two runs, by position, measured from the stored logits (\cref{sec:tokensweep}); ``prompt'' is the first generated token.
Colours and markers are shared between the panels and labelled in (b).}
\label{fig:acts}
\end{figure}

\subsection{Prefill: small, rounding-scale, structureless noise}
\label{sec:acts-prefill}

\Cref{fig:acts} and \cref{tab:acts} (\cref{app:acts}) summarise the prompt position.
Four regularities stand out.

\textbf{\fpt{} prefill is batch-invariant for Llama and Qwen on the A100.}
Changing $B$ leaves 99.9--100\% of their prefill vectors bitwise identical: the kernels chosen for these shapes evidently reduce in the same order.
Gemma on the L4 is not bitwise invariant, though its differences are at the level of \fpt{} rounding (median relative $\ell_2 \approx 5\times10^{-7}$, a few tens of ULPs).
Both comparisons hold the GPU fixed but differ in model and GPU at once, so we cannot say which explains the contrast; batch invariance is a property of a kernel set and a shape, not of the format.
Under \fpt{}, decode positions are no longer bitwise identical (median 0.02\% of rows for Llama and Qwen, none for Gemma), but the residual is sub-ULP in scale: median relative $\ell_2$ $7\times10^{-7}$, at most $6\times10^{-6}$ at the 99th percentile, with cosine similarity 1.000 to four decimals.
Bitwise identity therefore holds only at the prompt; at decode positions \fpt{} is reproducible to rounding, not to the bit.

\textbf{In \bft{}, batch-size noise is as large as the precision change itself, on the one model where we can separate them.}
For Gemma, where both sides run on the L4, the median relative $\ell_2$ between two \bft{} runs with different $B$ is $4.9\times10^{-3}$, slightly above the $4.4\times10^{-3}$ between \fpt{} and \bft{} at the same $B$: changing the batch size alone moves the representation further than the entire precision conversion.
The 8B models point the same way ($1.3\times10^{-2}$ vs.\ $1.0\times10^{-2}$ for Llama, $1.0\times10^{-2}$ vs.\ $0.85\times10^{-2}$ for Qwen), but their \fpt{} side also changes GPU, so we treat only the Gemma comparison as decisive.
That single clean comparison carries an 11\% margin on one model at one size, so we state it as scoped evidence rather than a general property of \bft{}; separating the two effects on a second model needs only a model whose \fpt{} weights fit on one device.
A probe trained on \bft{} activations at one batch size is thus already off-distribution, in this sense, at another.

\textbf{\fps{} is about $8\times$ quieter than \bft.}
The same batch-size change gives a median relative $\ell_2$ of $1.7\times10^{-3}$ (Llama) and $1.3\times10^{-3}$ (Qwen) in \fps{}, a ratio of 7.8 and 7.7 to \bft.
Both sides of every comparison here run on the L4, so this contrast is between formats alone.
This matches the three extra mantissa bits of \fps{} ($2^3=8$).
Measured in native units, both formats differ by 2--4 representable steps per element (\cref{tab:acts}).
The perturbation is therefore rounding noise at the resolution of the format, not a qualitative change in computation.

\textbf{The noise has no directional drift and is not amplified with depth.}
If batch size produced a systematic drift, distances would be additive, $d(4,16)\approx d(4,8)+d(8,16)$; instead every non-\fpt{} triple fits $d(4,16)\approx d(4,8)\approx d(8,16)$ (224/224; median relative error 1.5\% at prefill versus 50\% for the additive model), so each batch size is an independent draw of rounding noise.
The median per-block growth factor of prefill divergence is 1.01 and deeper blocks are only $1.6\times$ noisier than block 3 (range $1.1$--$6.4\times$), so there is no exponential amplification, and relative $\ell_2$ is nearly unrelated to input norm or sequence length (median $R^2=0.05$).

\textbf{Global geometry is preserved, but local neighbourhoods are not exact.}
Prefill CKA is at least 0.995 for every non-trivial comparison, yet 10-NN overlap is 0.95 under \bft{} (0.99 under \fps{}, minimum 0.70): noise at the $10^{-2}$ scale reorders near-tied neighbours, and the smaller the neighbourhood the less stable it is (5-NN 94.2\% versus 50-NN 95.9\% at the prompt, 43.5\% versus 50.6\% at token 20; \cref{app:geometry}).
A readout that depends on local structure, such as a $k$-NN or clustering-based monitor, is therefore more exposed than the probes we study.
Projected onto the data's principal components, the perturbation's variance ranks like the data's own (median Spearman 0.88), as multiplicative rounding error should, so the noise is not confined to directions a probe ignores.

\subsection{Decode: rounding noise becomes different text}
\label{sec:acts-decode}

During generation the situation changes qualitatively (\cref{fig:acts}b; \cref{tab:token} in \cref{app:token} gives the numbers).

\textbf{Rounding decides tokens.}
Under \fpt{}, changing the batch size leaves every realised token identical: agreement is exactly 1.000 at all four positions in all 18 pairs, and not one of 5{,}610 rows per model diverges anywhere, so the bitwise and sub-ULP results of \cref{sec:acts-prefill} carry through decoding.
Under \bft{}, the same change flips the \emph{first} generated token for 2.1\% of rows (1.5--3.0\% across models) and leaves the runs on different tokens in 25\% of rows by token 20; under \fps{} the figures are 0.4\% and 3.6\%.
Changing precision behaves like the noisier format involved (\fpt{} vs.\ \bft{} diverges somewhere in 32.6\% of rows, \fpt{} vs.\ \fps{} in 5.2\%).
The mechanism is thus observed rather than inferred: a perturbation at the scale of the format's rounding flips an argmax near-tie, and the model proceeds on a different sequence.

Divergence is not confined to late steps: 8.3\% of diverging rows already differ at the first generated token and 33.0\% by token 5, so a monitor reading even a short generation is exposed.

\textbf{The activation-distance criterion was accurate.}
Checked against the token ids it stood in for (\cref{fig:proxy}, \cref{app:fork}), our block-3 threshold agrees to a mean absolute difference of 0.95\pp{} over the 26 cells where both exist ($r=0.993$), with no systematic bias, so the analyses resting on it measure what they claim to.

Divergence also erodes population-level similarity: median CKA at token 20 falls to 0.64 (\bft, batch size) and kNN overlap to 0.45 (\cref{tab:acts}), extending the output-level observations of \citet{yuan2025nondeterminism} to internal states.

% ---------------------------------------------------------------------------
\section{How much do probes change across configurations?}
\label{sec:probes}

\paragraph{Prefill probes are effectively configuration-invariant.}
Across the 5{,}568 transfer measurements (summarised in \cref{fig:probe-delta}, in full in \cref{app:transfer}), at the last prompt token 74\% of transfers leave accuracy exactly unchanged, mean $|\Delta|$ is 0.04\pp{} and $|\Delta|$ never exceeds 0.47\pp{} across all 1{,}392 cells, with gains and losses balanced (178 of 365 non-zero changes negative, binomial $p=0.68$).
Transfers involving \bft{} change most and those between \fpt{} and \fps{} are nearly exact (at most 0.02\pp; \cref{fig:heatmap}), while probes trained on \fpt{} are exactly unchanged under any batch-size change at every position (576/576 cells) -- from bitwise identity at the prompt for Llama and Qwen, and from sub-ULP differences elsewhere.
Numerical noise at the $10^{-2}$ scale thus shifts the representation far less than the class margin of any of our probes, even though it does not avoid the high-variance directions (\cref{sec:acts-prefill}).

\paragraph{Decode probes change in most cells, in proportion to divergence.}
At generated tokens 5, 10 and 20, accuracy changes in 69\%, 75\% and 80\% of cells, with mean $|\Delta|$ 0.19, 0.28 and 0.37\pp{} and a maximum of 2.35\pp; 4.4\% of decode cells exceed 1\pp.
The changes are nearly balanced (mean $\Delta=-0.02$\pp), so transfer reshuffles which examples are correct rather than degrading steadily, and \cref{fig:probe-delta}b links this to activations: excluding cells that are zero on both axes by construction, mean $|\Delta|$ rises with the diverged fraction from 0.03\pp{} to 0.49\pp{} above 20\%.
The diverged fraction predicts $|\Delta|$ with Spearman $\rho=0.73$, and \cref{fig:probe-delta}c recomputes that within subsets: it is not carried by the hardware confound ($0.76$ same-GPU, $0.65$ cross-GPU), nor by precision standing in for both quantities ($0.53$ within \bft{} and $0.54$ within \fps{} with the format held fixed), and the strictest subset -- decode positions, same GPU, construction-zero cells dropped -- still gives $\rho=0.47$ ($n=864$, $p<10^{-48}$).

\begin{table}[t]
\caption{Per-example agreement between the two runs of a pair.
Each table row pools 174 run pairs $\times$ 3 probe types $\times$ the dataset's test split (1{,}066 or 804 examples) $=488{,}070$ row evaluations, of which 387{,}690 are ``same tokens'' and 100{,}380 ``diverged''; the \fpt{} batch-size pairs quoted in the text are 50{,}490 of these per position, or 201{,}960 across all four.
``Verdicts that flip'' is $P[\hat y_A \neq \hat y_B]$; $|\Delta$acc$|$ is computed over \emph{exactly the same rows}, so the gap between those columns is not a difference of sample but what aggregate accuracy conceals.
$\kappa$ is Cohen's kappa between the runs' predictions, score $r$ the correlation of their class-1 probabilities.}
\label{tab:agreement}
\centering
\small
\begin{tabular}{lrrrrrrr}
\toprule
& \multicolumn{3}{c}{Verdicts that flip (\%)} & & & & \\
\cmidrule(lr){2-4}
Probe reads & all rows & same tokens & diverged & $|\Delta$acc$|$ (pp) & $\kappa$ & $\Delta$AUROC (pp) & score $r$ \\
\midrule
prompt & 0.08 & 0.07 & 0.11 & 0.04 & 0.999 & +0.00 & 1.000 \\
gen 5 & 0.87 & 0.12 & 3.75 & 0.18 & 0.981 & +0.02 & 0.990 \\
gen 10 & 1.64 & 0.13 & 7.48 & 0.23 & 0.965 & -0.01 & 0.978 \\
gen 20 & 2.76 & 0.15 & 12.87 & 0.31 & 0.941 & +0.05 & 0.961 \\
\bottomrule
\end{tabular}

\end{table}

\paragraph{Reading a diverged token costs the probe nothing, so why do decode transfers move more?}
Evaluating each probe on run B's activations, split by whether that row's trajectory matched run A's, the accuracy difference is $+0.24$\pp{} at the prompt and $-0.17$, $-0.05$ and $-0.11$\pp{} at tokens 5, 10 and 20, changing sign and never exceeding $0.58$\pp{} in the mean over cells (\cref{app:transfer}): diverged text is not harder for the probe to read.
What does change is the margin: accuracy on identical-trajectory rows falls from 94.6\% at the prompt to 91.7\%, 89.5\% and 87.7\%, so a probe reading a later token starts closer to its boundary.
The larger decode-position changes therefore combine that smaller margin with a differing input, not a probe mishandling diverged text.

\begin{figure}[t]
\centering
\includegraphics[width=0.60\linewidth]{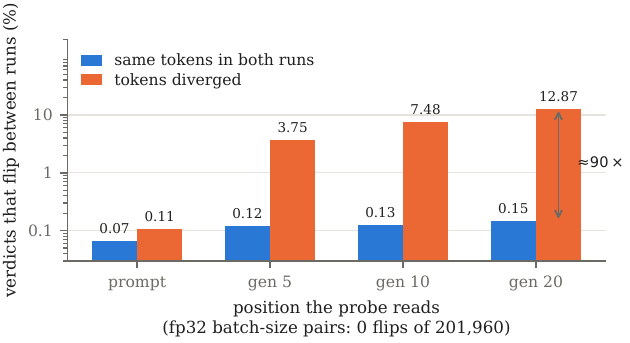}
\caption{How often a probe's verdict changes between two runs of the same model and dataset, by the position the probe reads, pooled over 174 run pairs and three probe types.
Rows are split by whether the two runs' realised tokens matched at every logged position: the same-token rate is flat at 0.12--0.15\% across decode positions while the diverged rate quadruples, so what moves a verdict is the text changing, not the arithmetic.
A log axis cannot render an exact zero, so note that the \fpt{} batch-size pairs are included in these bars and contribute no flips at all.}
\label{fig:flips}
\end{figure}

\paragraph{How often does an individual verdict change?}
Accuracy, however, is the wrong instrument for a monitor: it is invariant to \emph{which} examples are correct, so everything above is consistent with a probe that quietly changes thousands of individual decisions.
Running each probe over \emph{both} runs' activations for the same rows measures what matters, and the two differ by up to an order of magnitude: at token 20 accuracy moves 0.31\pp{} while 2.76\% of verdicts change (\cref{tab:agreement}), a ratio widening from $2\times$ at the prompt to $9\times$.
\Cref{fig:flips} shows where those flips come from, and the answer is the text rather than the arithmetic.
At the prompt, where both runs read identical input, 0.076\% of verdicts change: noise at the $10^{-2}$ scale almost never moves a decision.
At decode positions the rate rises to 2.76\%, but conditioning on trajectory splits it cleanly, and the same-token rate stays flat at 0.12--0.15\% while the diverged rate quadruples to 12.9\%.
That residual is twice the prompt rate, too consistent to be chance ($p<10^{-10}$) though tiny, and some is expected since a row counts as ``same tokens'' only at the four logged positions (\cref{sec:tokensweep}).
Among diverged rows the flip rate is nearly constant across perturbations (12.5--15.8\%, \cref{tab:agreement-pert}): a perturbation decides how often the text changes, not what changed text does to the probe.

\paragraph{Flips are symmetric, and ranking survives.}
The changes favour neither class: at token 20, 6{,}671 verdicts move from class 0 to 1 against 6{,}813 the other way, and the flip rate on positives (2.65\%) matches that on negatives (2.84\%), so a monitor gains about as many false alarms as it loses detections.
Threshold-free behaviour is steadier still: $\kappa$ is 0.999 at the prompt and 0.941 at token 20, scores correlate at $r=1.000$ and $0.961$, and AUROC moves by at most 0.05\pp{} (TPR at 1\% FPR by at most 0.34\pp), so what is disturbed is the handful of examples within rounding distance of the boundary, not the ordering of the rest.
Under \fpt{} with only the batch size changed, not one verdict flips in any of the 201{,}960 evaluations those 18 pairs contribute across the four positions ($\kappa=1.000$), while the prompt flip rate otherwise tracks the mantissa: 0.14\% under \bft{} against 0.018\% under \fps.

\paragraph{The model's own decision is the fragile one.}
Both quantities come from the same prompt activations under the same perturbation: a \bft{} batch-size change flips 0.14\% of probe verdicts but 2.1\% of the model's own next-token argmaxes, a $15\times$ difference.
The reason is margin: a binary probe with a well-separated boundary absorbs a $10^{-2}$ relative perturbation, whereas an argmax over $10^{5}$ tokens often has a near-tie at the top.
Probes are robust not because the representation is stable in absolute terms, but because their decision needs much less of it to be right.

\paragraph{The GPU change does not drive these results.}
\Cref{tab:gpu-split} splits every transfer by whether the training and test activations came from the same GPU.
The two classes behave alike: at the prompt, same-GPU transfers change accuracy by at most 0.47\pp{} and cross-GPU ones by at most 0.38\pp; at decode positions they change 72\% and 78\% of cells, and the per-example flip rate is equally unmoved (0.088\% versus 0.076\% at the prompt).
Restricting the precision axis to the 576 transfers that hold the GPU fixed leaves every conclusion unchanged, but we draw only the weaker of the two available conclusions from it.
The same-GPU and cross-GPU sets differ in composition as well as in hardware: every same-GPU precision transfer involves \bft{}, the noisier format, whereas about half the cross-GPU ones are \fps{}-only, so the two means cannot be differenced to estimate a GPU effect.
What \cref{tab:gpu-split} does establish is that the same-GPU subset alone supports every conclusion we draw; the independent contribution of the GPU is not identified by this design.

% ---------------------------------------------------------------------------
\section{Discussion}
\label{sec:discussion}

\paragraph{Implications for probe deployment.}
Probes reading the prompt, as in input classifiers, pre-generation monitors and most truthfulness probes, can be trained under one configuration and served under another: the worst case here is a 0.47\pp{} accuracy change and one verdict in 1{,}300 flipping.
Probes reading generated tokens face a different problem, since the text they judge depends on the serving configuration; such deployments should use batch-invariant decoding \citep{he2025nondeterminism,lmsys2025deterministic}, re-encode the generated text deterministically before probing, or measure their own flip rate.
For auditing, \fpt{} is exactly reproducible across batch sizes here: in prompt activations, in every realised token and in every probe verdict.

\paragraph{Three regimes, and what aggregate accuracy hides.}
The three regimes named in \cref{sec:intro} are one each here: \emph{benign} at the prompt; \emph{behaviourally consequential} during decoding, changing a discrete choice and hence the model's future input; and a \emph{validity failure} for Gemma-3-4B in \fps{} (\cref{sec:gemma}).
Ignore the first, measure the second under the serving configuration, detect and refuse the third.
Accuracy cannot tell the first two apart, being invariant to which examples are right: it understates the verdict churn by $2$--$9\times$, so a monitor judged on accuracy alone looks stable while changing thousands of decisions.

\paragraph{Limitations.}
(i) One software stack, and one GPU except that the 8B models' \fpt{} runs used an A100, so for those models precision and GPU move together; the same-GPU subset supports every conclusion (\cref{tab:gpu-split}), but the GPU's own contribution is not identified, and \fpt{} batch invariance is a property of one GPU and kernel set.
(ii) Batch sizes 4--16 are small for production serving; larger, ragged or continuously batched workloads and tensor parallelism may select other kernels.
(iii) Two binary tasks (804 and 1{,}066 test examples); safety probes run at low false-positive rates on harder distributions, where margins are smaller, and should be more sensitive.
(iv) Per-example agreement and its score-level companions are measured on the test split only, at the final layer, for an \fpt{}-trained probe; the wider transfer grid reports accuracy alone.
(v) Each probe is trained once, so we lack a seed baseline for judging a 0.47\pp{} change; the control in \cref{sec:setup} fixes the seed and supplies none.
(vi) The token sweep sees only the four logged positions, so a divergence that self-corrects between two of them is invisible; this is the likeliest source of the residual 0.12--0.15\% flip rate on rows counted as matching.
(vii) We study 4B--8B dense models; quantized weights \citep{singh2025quantization} and larger models are natural next steps.

% ---------------------------------------------------------------------------
\section{Conclusion}
\label{sec:conclusion}

Batch size and precision perturb LLM activations at the scale of the format's rounding, and at the prompt this does not matter for probes: across three models, two tasks and three architectures, transfer moves accuracy by under half a point and flips under one verdict in a thousand.
During generation the same noise flips greedy token choices, and a probe's verdict changes almost only on rows whose tokens changed: the probe is robust and its input is not.
Hence the recommendation we would most like carried forward: \emph{evaluate an activation monitor on per-example agreement, not aggregate accuracy}, which hid the churn here by two to nine times.

% ---------------------------------------------------------------------------
\subsection*{AI use statement}
Generative AI tools were used to help draft and edit the text of this paper, to write analysis and plotting scripts that aggregate our stored results, and to search for and verify related literature.
All experimental design, data collection, and scientific claims were checked by the authors, who take full responsibility for the content.

\subsection*{Ethics statement}
This work measures the reliability of activation probes, which are used as safety monitors.
It uses public models and public datasets and involves no human subjects.
We see no direct negative use; our findings may help practitioners avoid over- or under-estimating the stability of deployed monitors.

\subsection*{Reproducibility statement}
\Cref{sec:setup} specifies models, datasets, splits, the configuration grid, extracted layers and positions, probe architectures and hyperparameters; \cref{app:metrics} defines every activation metric, \cref{app:fork} the divergence criterion, and \cref{app:followup} what the stored data can still answer.
\Cref{app:agreement,app:token,app:acts,app:transfer,app:gpu,app:geometry,app:matched} report the full per-model and per-configuration breakdowns behind every pooled number in the main text.
Dataset manifests are pinned to fixed repository revisions, splits are a deterministic function of row content, and all runs use seed 42 with deterministic PyTorch algorithms.
We will release extraction, probe-training and analysis code, the per-row metric caches, and the scripts that generate every table and figure in this paper, with the \fps{} Gemma-3-4B exclusion applied in code.

\bibliography{refs}
\bibliographystyle{iclr2027_conference}

% ---------------------------------------------------------------------------
\appendix

\section{Metric details}
\label{app:metrics}

\paragraph{ULP distance.}
For \fpt{} arrays we reinterpret each value's bit pattern as a monotone integer ordering and take the absolute integer difference.
This counts how many representable \fpt{} numbers lie between two elements.
Values from \bft{} and \fps{} runs are upcast losslessly, so one step of their native grid spans $2^{16}$ (\bft) or $2^{13}$ (\fps) \fpt{} steps for values of matching exponent.

\paragraph{Triangle test.}
For each model, dataset, precision, layer and position, we compare $d(4,16)$ against an additive prediction $d(4,8)+d(8,16)$ and a common-mean prediction $\tfrac12(d(4,8)+d(8,16))$, where $d$ is the mean per-row relative $\ell_2$.
We report which prediction is closer.

\paragraph{Per-block amplification.}
For consecutive recorded blocks $\ell_1<\ell_2$ we report $(d_{\ell_2}/d_{\ell_1})^{1/(\ell_2-\ell_1)}$, using block 3 as the floor.

\paragraph{Geometry.}
Linear CKA and orthogonal-Procrustes residual (relative to $\|\mX_a\|_F$) use centred matrices of up to 3{,}000 rows.
kNN overlap is the mean Jaccard index of each row's 10 nearest Euclidean neighbours in the two runs (1{,}500 rows).
The spectral-alignment statistic is the Spearman rank correlation between the data variance and the perturbation variance along the top 50 principal components of run $a$.

\paragraph{Input dependence.}
We regress per-row relative $\ell_2$ on $\|\vx_a\|_2$ and sequence length by OLS.

\section{Bimodality of decode-position divergence}
\label{app:fork}

Pooling \bft{} batch-size comparisons at block 3 over all models and datasets, per-row relative $\ell_2$ at the prompt lies entirely in $[10^{-3},10^{-1.5})$.
The only exceptions are 24 rows (0.04\%) with distance exactly zero.
At token 20, 74\% of rows remain in $[10^{-4.5},10^{-1.5})$, a second mode in $[10^{-0.5},10^{0.5})$ holds 24\%, and only 1.5\% fall in the valley $[10^{-1.5},10^{-0.5})$ between them.
The 0.1 threshold lies inside this valley, so any threshold in $[10^{-1.5},10^{-0.5}]$ changes this pooled diverged fraction by at most 1.5\pp.

\begin{figure}[H]
\centering
\includegraphics[width=0.46\linewidth]{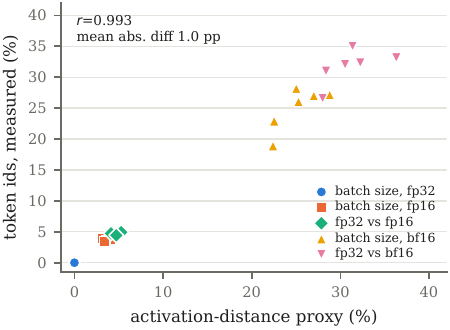}
\caption{The activation-distance criterion of \cref{sec:setup} against the token-id measurement it stood in for, one point per (model, dataset, perturbation) at generated token 20.
The dotted line is equality.}
\label{fig:proxy}
\end{figure}

\section{What the stored data can still answer}
\label{app:followup}

\textbf{Done: trajectory reconstruction.}
\texttt{SAVE\_LOGITS} was on during extraction, so every logged position of every run has its output logit vector on disk, and under greedy decoding the token fed to the next step is that vector's argmax.
\Cref{sec:tokensweep,sec:acts-decode} report the resulting comparison: per-position token agreement, the first logged position at which two runs disagree, and probe accuracy on the identical- and diverged-trajectory subsets.
The one irreducible gap is resolution, since only positions 5, 10 and 20 were logged.

\textbf{Done: probes at decode positions.}
The sweep evaluates each probe at all four logged positions, so it compares a probe reading an activation produced by a token the other run never generated against one reading an agreed token (\cref{sec:probes}).
The gaps are centred on zero at every position, which is what attributes the decode-position transfer changes to the input differing rather than to the probe mishandling diverged text.

\textbf{Done: per-example agreement.}
Running each probe over both runs' activations for the same rows yields $P[\hat y_A \neq \hat y_B]$, its split by flip direction and by true label, Cohen's $\kappa$, and score-level metrics, reported in \cref{sec:probes,tab:agreement} and broken down by perturbation in \cref{tab:agreement-pert}.

\textbf{Outstanding: seed variance.}
Retraining a subset of probes under several seeds would give the noise floor against which a 0.47\pp{} accuracy change, or a 0.076\% prompt flip rate, should be read.
That is the one remaining measurement, and it needs probe training rather than new inference.

\section{Verdict flips by perturbation}
\label{app:agreement}

\begin{table}[H]
\caption{Percentage of individual probe verdicts that change between the two runs, by perturbation, pooled over models, datasets and probe types.
The last two columns split generated token 20 by whether that row's realised tokens matched.
\fpt{} batch-size pairs flip no verdict anywhere; a diverged subset that no row enters is marked ``--''.
Note the last two columns: the flip rate among diverged rows is nearly constant across perturbations (12.5--15.8\%), so what the perturbation controls is how often the text changes, not what a changed text does to the probe.}
\label{tab:agreement-pert}
\centering
\small
\begin{tabular}{lrrrrrr}
\toprule
& \multicolumn{4}{c}{Verdicts that flip, all rows (\%)} & \multicolumn{2}{c}{At gen 20 (\%)} \\
\cmidrule(lr){2-5}\cmidrule(lr){6-7}
Perturbation & prompt & gen 5 & gen 10 & gen 20 & same tokens & diverged \\
\midrule
Batch size (fp32) & 0.000 & 0.000 & 0.000 & 0.000 & 0.000 & -- \\
Batch size (fp16) & 0.018 & 0.160 & 0.267 & 0.630 & 0.040 & 15.756 \\
fp32 vs fp16 & 0.012 & 0.160 & 0.371 & 0.719 & 0.028 & 13.345 \\
Batch size (bf16) & 0.143 & 1.121 & 2.183 & 3.628 & 0.283 & 13.190 \\
fp32 vs bf16 & 0.117 & 1.382 & 2.686 & 4.371 & 0.235 & 12.877 \\
bf16 vs fp16 & 0.102 & 1.342 & 2.348 & 4.053 & 0.255 & 12.532 \\
\bottomrule
\end{tabular}

\end{table}

\section{Token divergence in full}
\label{app:token}

\begin{table}[H]
\caption{Greedy token disagreement between two runs, measured from stored logits on the test split.
``First token'' is the argmax at the last prompt position, ``any position'' the share of rows disagreeing somewhere.
Models are Gemma-3-4B (G), Llama-3.1-8B (L) and Qwen3-8B (Q); the range is over model--dataset pairs.
The probe gap is the mean accuracy difference (identical minus diverged rows) for a probe reading the prompt and one reading generated token 20.
Per-model gaps scatter in both directions and should not be read individually: the \fps{} rows rest on a diverged subset of roughly 40 examples, and even where that subset is large (Gemma at 26.5\% diverged, $+2.59$\pp) a single model--dataset cell carries wide sampling error.
The pooled estimate in \cref{sec:probes}, which is centred on zero at every position, is the one to read.
Gemma-3-4B contributes no \fps{} comparisons (\cref{sec:gemma}); \cref{tab:token-full} breaks this table down by model.}
\label{tab:token}
\centering
\small
\resizebox{\linewidth}{!}{\begin{tabular}{llrrrrrrr}
\toprule
& & & \multicolumn{4}{c}{Greedy token disagreement (\%)} & \multicolumn{2}{c}{Probe gap (pp)} \\
\cmidrule(lr){4-7}\cmidrule(lr){8-9}
Perturbation & Models & pairs & first token & at token 20 & any position & range & prompt & token 20 \\
\midrule
Batch size (fp32) & G, L, Q & 18 & 0.00 & 0.0 & 0.0 & [0.0, 0.0] & -- & -- \\
Batch size (fp16) & L, Q & 12 & 0.36 & 3.6 & 3.7 & [3.0, 4.4] & +0.90 & +1.96 \\
fp32 vs fp16 & L, Q & 36 & 0.36 & 4.7 & 5.2 & [4.7, 6.1] & +1.18 & +1.85 \\
Batch size (bf16) & G, L, Q & 18 & 2.05 & 25.1 & 25.8 & [19.3, 30.5] & +0.67 & +0.11 \\
fp32 vs bf16 & G, L, Q & 54 & 2.57 & 31.7 & 32.6 & [26.5, 36.7] & +0.13 & -0.21 \\
bf16 vs fp16 & L, Q & 36 & 2.85 & 29.7 & 30.9 & [25.5, 35.0] & -0.73 & -1.10 \\
\bottomrule
\end{tabular}
}
\end{table}

\begin{table}[H]
\caption{\Cref{tab:token} broken down by model, with row counts. ``Rows'' counts (row, pair) comparisons.}
\label{tab:token-full}
\centering
\small
\resizebox{\linewidth}{!}{\begin{tabular}{llrrrrrr}
\toprule
& & & \multicolumn{3}{c}{Greedy token disagreement (\%)} & \multicolumn{2}{c}{Probe gap (pp)} \\
\cmidrule(lr){4-6}\cmidrule(lr){7-8}
Perturbation & Model & rows & first token & at token 20 & any position & prompt & token 20 \\
\midrule
Batch size (fp32) & Llama-3.1-8B & 5610 & 0.00 & 0.0 & 0.0 & -- & -- \\
Batch size (fp32) & Qwen3-8B & 5610 & 0.00 & 0.0 & 0.0 & -- & -- \\
Batch size (fp32) & Gemma-3-4B & 5610 & 0.00 & 0.0 & 0.0 & -- & -- \\
\midrule
Batch size (fp16) & Llama-3.1-8B & 5610 & 0.41 & 3.7 & 3.8 & +0.22 & -0.57 \\
Batch size (fp16) & Qwen3-8B & 5610 & 0.31 & 3.6 & 3.6 & +1.58 & +4.50 \\
\midrule
fp32 vs fp16 & Llama-3.1-8B & 16830 & 0.38 & 4.7 & 5.1 & +0.24 & -0.68 \\
fp32 vs fp16 & Qwen3-8B & 16830 & 0.34 & 4.7 & 5.2 & +2.11 & +4.38 \\
\midrule
Batch size (bf16) & Llama-3.1-8B & 5610 & 1.54 & 20.9 & 21.8 & -0.03 & +0.93 \\
Batch size (bf16) & Qwen3-8B & 5610 & 3.00 & 27.7 & 28.5 & -0.56 & -1.92 \\
Batch size (bf16) & Gemma-3-4B & 5610 & 1.62 & 26.5 & 27.2 & +2.59 & +1.32 \\
\midrule
fp32 vs bf16 & Llama-3.1-8B & 16830 & 2.40 & 29.3 & 30.5 & -0.28 & -0.25 \\
fp32 vs bf16 & Qwen3-8B & 16830 & 3.35 & 34.1 & 35.1 & -0.57 & -1.47 \\
fp32 vs bf16 & Gemma-3-4B & 16830 & 1.96 & 31.6 & 32.2 & +1.23 & +1.09 \\
\midrule
bf16 vs fp16 & Llama-3.1-8B & 16830 & 2.24 & 27.6 & 28.7 & -0.53 & -0.17 \\
bf16 vs fp16 & Qwen3-8B & 16830 & 3.45 & 31.8 & 33.0 & -0.94 & -2.02 \\
\bottomrule
\end{tabular}
}
\end{table}

\section{Activation divergence in full}
\label{app:acts}

\begin{table}[H]
\caption{Activation divergence by perturbation type.
The GPU column gives the hardware on the two sides of the comparison; \texttt{A100/L4} rows change GPU and precision together (\cref{sec:setup}).
Prefill columns are medians over layers, datasets and pairs.
kNN$_{10}$ is the 10-nearest-neighbour Jaccard overlap at decode token 20.
``Diverged'' is the percentage of rows with block-3 relative $\ell_2>0.1$ at token 20, i.e.\ processing a different token there.
ULP counts the \fpt{} values between two elements.
One native \bft{} step equals $2^{16}$ \fpt{} ULPs and one \fps{} step equals $2^{13}$, so the typical element differs by 2--4 native steps under either format.}
\label{tab:acts}
\centering
\small
\resizebox{\linewidth}{!}{\begin{tabular}{lllrrrrrr}
\toprule
& & & \multicolumn{4}{c}{Prefill (last prompt token)} & \multicolumn{2}{c}{Decode (token 20)} \\
\cmidrule(lr){4-7}\cmidrule(lr){8-9}
Perturbation & Model & GPU(s) & bit-id.\ (\%) & med.\ rel.\ $\ell_2$ & med.\ ULP & CKA & kNN$_{10}$ & diverged (\%) \\
\midrule
Batch size (fp32) & Llama-3.1-8B & A100 & 100.0 & 0 & 0 & 1.0000 & 1.00 & 0.0 \\
Batch size (fp32) & Qwen3-8B & A100 & 100.0 & 0 & 0 & 1.0000 & 1.00 & 0.0 \\
Batch size (fp32) & Gemma-3-4B & L4 & 0.0 & $4.8\!\times\!10^{-7}$ & 33 & 1.0000 & 1.00 & 0.0 \\
\midrule
Batch size (fp16) & Llama-3.1-8B & L4 & 0.0 & $1.7\!\times\!10^{-3}$ & 24,576 & 1.0000 & 0.89 & 3.3 \\
Batch size (fp16) & Qwen3-8B & L4 & 0.0 & $1.3\!\times\!10^{-3}$ & 24,576 & 1.0000 & 0.86 & 4.0 \\
\midrule
fp32 $\to$ fp16 & Llama-3.1-8B & L4/A100 & 0.0 & $1.6\!\times\!10^{-3}$ & 21,004 & 1.0000 & 0.86 & 4.6 \\
fp32 $\to$ fp16 & Qwen3-8B & L4/A100 & 0.0 & $1.1\!\times\!10^{-3}$ & 22,096 & 1.0000 & 0.82 & 4.7 \\
\midrule
Batch size (bf16) & Llama-3.1-8B & L4 & 0.0 & $1.3\!\times\!10^{-2}$ & 196,608 & 0.9999 & 0.49 & 21.6 \\
Batch size (bf16) & Qwen3-8B & L4 & 0.0 & $1.0\!\times\!10^{-2}$ & 196,608 & 0.9999 & 0.43 & 25.1 \\
Batch size (bf16) & Gemma-3-4B & L4 & 0.0 & $4.9\!\times\!10^{-3}$ & 262,144 & 0.9988 & 0.40 & 26.1 \\
\midrule
fp32 $\to$ bf16 & Llama-3.1-8B & L4/A100 & 0.0 & $1.0\!\times\!10^{-2}$ & 138,859 & 0.9999 & 0.42 & 29.5 \\
fp32 $\to$ bf16 & Qwen3-8B & L4/A100 & 0.0 & $8.5\!\times\!10^{-3}$ & 169,468 & 0.9999 & 0.36 & 32.1 \\
fp32 $\to$ bf16 & Gemma-3-4B & L4 & 0.0 & $4.4\!\times\!10^{-3}$ & 216,277 & 0.9992 & 0.35 & 30.3 \\
\bottomrule
\end{tabular}
}
\end{table}

\section{Probe transfer in full}
\label{app:transfer}

\begin{figure}[H]
\centering
\includegraphics[width=\linewidth]{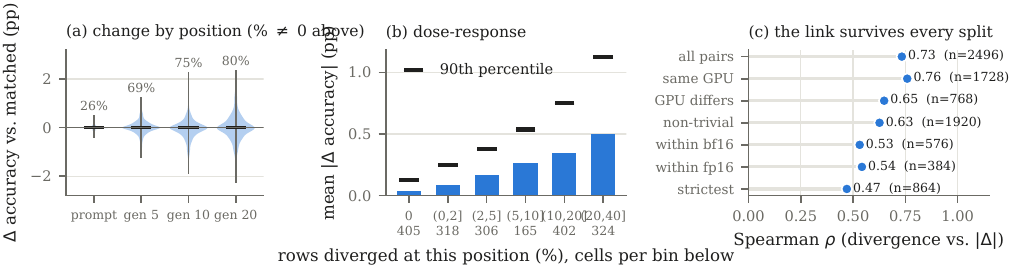}
\caption{\textbf{(a)} Distribution of accuracy change $\Delta$ over all 5{,}568 transfer cells, by probed position; the labels give the share of cells with $\Delta\neq0$.
\textbf{(b)} Mean $|\Delta|$ (bars) and 90th percentile (ticks), grouped by the percentage of rows that diverged in the corresponding activation comparison, over the 2{,}496 transfer cells whose train/test pair differs only in batch size, or in precision with \fpt{} on one side.
The 576 cells that are \fpt{} on both sides and differ only in batch size are zero on both axes by construction and are excluded here.
\textbf{(c)} The same rank correlation recomputed within subsets, to check that it is not an artefact of the hardware confound, of precision standing in for both quantities, or of the construction-zero cells.}
\label{fig:probe-delta}
\end{figure}

Secondary factors, pooled over the cells of \cref{tab:probe-transfer}.
Probe architecture has a modest effect at decode positions: LR changes in 78\% of cells (mean $|\Delta|$ 0.33\pp) versus 71\% for MLP-1 (0.24\pp) and 74\% for MLP-3 (0.28\pp), so capacity neither buys nor costs robustness here.
The harder task is more sensitive (mean decode $|\Delta|$ 0.36\pp{} on Rotten Tomatoes versus 0.20\pp{} on SMS Spam), and depth matters less than position (0.18--0.28\pp{} across blocks, against 0.04--0.37\pp{} across positions).

\begin{table}[H]
\caption{Probe transfer across inference configurations: $\Delta$ is the change in test accuracy (percentage points) relative to the probe's matched configuration, pooled over models, datasets, layers and probe types.
``Changed'' is the percentage of cells with $\Delta\neq0$.
Training batch size is always 4.
Rows that change precision pool same-GPU and cross-GPU transfers; \cref{tab:gpu-split} separates them.
Gemma-3-4B contributes no \fps{} cells (\cref{sec:gemma}).}
\label{tab:probe-transfer}
\centering
\small
\resizebox{\linewidth}{!}{\begin{tabular}{lllrrrrrr}
\toprule
Stage & Train & Test differs in & cells & changed (\%) & mean $|\Delta|$ & max $|\Delta|$ & $|\Delta|{>}1$pp (\%) & mean $\Delta$ \\
\midrule
prefill & fp32 & batch size only & 144 & 0 & 0.00 & 0.00 & 0.0 & +0.00 \\
 &  & precision only & 120 & 28 & 0.04 & 0.38 & 0.0 & +0.00 \\
 &  & both & 240 & 29 & 0.04 & 0.38 & 0.0 & +0.01 \\
\cmidrule(lr){2-9}
 & fp16 & batch size only & 96 & 7 & 0.01 & 0.19 & 0.0 & -0.00 \\
 &  & precision only & 96 & 23 & 0.03 & 0.28 & 0.0 & +0.01 \\
 &  & both & 192 & 23 & 0.03 & 0.47 & 0.0 & +0.01 \\
\cmidrule(lr){2-9}
 & bf16 & batch size only & 144 & 42 & 0.06 & 0.38 & 0.0 & +0.00 \\
 &  & precision only & 120 & 35 & 0.05 & 0.38 & 0.0 & -0.01 \\
 &  & both & 240 & 35 & 0.05 & 0.38 & 0.0 & -0.01 \\
\midrule
decode & fp32 & batch size only & 432 & 0 & 0.00 & 0.00 & 0.0 & +0.00 \\
 &  & precision only & 360 & 79 & 0.29 & 1.59 & 4.7 & -0.01 \\
 &  & both & 720 & 79 & 0.29 & 2.35 & 5.4 & +0.03 \\
\cmidrule(lr){2-9}
 & fp16 & batch size only & 288 & 65 & 0.13 & 0.84 & 0.0 & -0.02 \\
 &  & precision only & 288 & 83 & 0.27 & 1.49 & 2.4 & -0.01 \\
 &  & both & 576 & 81 & 0.28 & 2.25 & 3.8 & +0.01 \\
\cmidrule(lr){2-9}
 & bf16 & batch size only & 432 & 89 & 0.39 & 1.78 & 7.9 & -0.00 \\
 &  & precision only & 360 & 90 & 0.38 & 1.97 & 5.8 & -0.08 \\
 &  & both & 720 & 90 & 0.38 & 1.97 & 5.8 & -0.09 \\
\bottomrule
\end{tabular}
}
\end{table}

\begin{figure}[H]
\centering
\includegraphics[width=\linewidth]{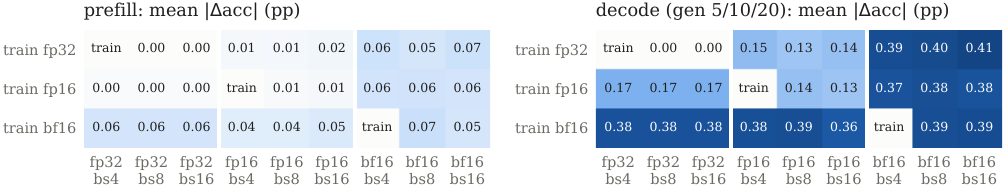}
\caption{Mean $|\Delta|$ (pp) for each training precision (rows) and test configuration (columns), pooled over models, datasets, layers and probes.
Training uses $B=4$.
Gemma-3-4B contributes only to \fpt{} and \bft{} cells.}
\label{fig:heatmap}
\end{figure}

\section{Transfers split by extraction GPU}
\label{app:gpu}

\begin{table}[H]
\caption{Probe transfer split by whether the training and test activations were extracted on the same GPU.
Cross-GPU cells are the \fpt{}-versus-\bft{}/\fps{} transfers of the two 8B models (\cref{sec:setup}); no batch-size-only transfer changes GPU.
$\Delta$ is in percentage points.}
\label{tab:gpu-split}
\centering
\small
\begin{tabular}{lllrrrrr}
\toprule
Stage & Extraction GPU & Test differs in & cells & changed (\%) & mean $|\Delta|$ & max $|\Delta|$ & mean $\Delta$ \\
\midrule
prefill & same GPU & batch size only & 384 & 17 & 0.02 & 0.38 & +0.00 \\
 &  & precision only & 144 & 42 & 0.06 & 0.38 & +0.00 \\
 &  & both & 288 & 41 & 0.06 & 0.47 & +0.01 \\
\cmidrule(lr){2-8}
 & GPU also differs & precision only & 192 & 20 & 0.03 & 0.28 & -0.00 \\
 &  & both & 384 & 22 & 0.03 & 0.38 & -0.01 \\
\midrule
decode & same GPU & batch size only & 1152 & 50 & 0.18 & 1.78 & -0.01 \\
 &  & precision only & 432 & 91 & 0.38 & 1.97 & -0.04 \\
 &  & both & 864 & 91 & 0.39 & 2.25 & -0.02 \\
\cmidrule(lr){2-8}
 & GPU also differs & precision only & 576 & 79 & 0.27 & 1.88 & -0.03 \\
 &  & both & 1152 & 78 & 0.26 & 2.35 & -0.02 \\
\bottomrule
\end{tabular}

\end{table}

\section{Representational geometry}
\label{app:geometry}

\begin{figure}[H]
\centering
\includegraphics[width=\linewidth]{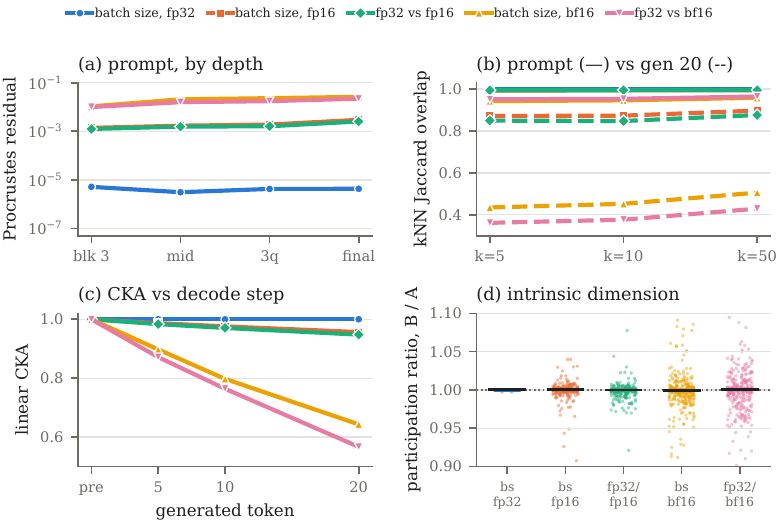}
\caption{Representational geometry under each perturbation, median over models, datasets, layers and datasets as applicable.
\textbf{(a)} Procrustes residual at the prompt by depth.
\textbf{(b)} kNN Jaccard overlap against neighbourhood size $k$, at the prompt (solid) and decode token 20 (dashed).
\textbf{(c)} Linear CKA against generated-token index.
\textbf{(d)} Ratio of the two runs' participation ratios; each point is one (model, dataset, layer, position, pair), bars are medians.
\fpt{}-versus-\bft{}/\fps{} comparisons change GPU as well as precision for the 8B models (\cref{sec:setup}).}
\label{fig:geometry}
\end{figure}

\Cref{fig:geometry} collects the geometry measurements behind \cref{tab:acts}.
Four observations are worth recording.

\textbf{(a) Rigid misalignment grows mildly with depth and tracks the format.}
The orthogonal-Procrustes residual at the prompt is $1.6\times10^{-2}$ for a \bft{} batch-size change and $1.8\times10^{-3}$ in \fps{}, a ratio of 9, close to the $8\times$ of the per-row distances.
It grows by a median $2.1\times$ from block 3 to the final block, against $1.5\times$ for the per-row relative $\ell_2$ over the same span.
The two differ because the Procrustes residual is a whole-matrix quantity, dominated by the few large-norm channels that grow with depth, whereas relative $\ell_2$ is normalised per row.
Both are small constant factors rather than the geometric growth that would follow from per-block amplification, so they tell the same story as \cref{sec:acts-prefill}.
Under \fpt{} it stays near $4\times10^{-6}$.

\textbf{(b) Small neighbourhoods are the least stable part of the geometry.}
At the prompt, \bft{} batch-size changes preserve 94.2\% of 5-nearest-neighbour sets but 95.9\% of 50-nearest-neighbour sets; at decode token 20 the same ordering is much starker (43.5\% versus 50.6\%).
Rounding noise reorders near-ties, which affects the tightest neighbourhoods first.
Probes that rely on local neighbourhood structure, such as $k$-NN readouts, are therefore more exposed than the linear and MLP probes we study.

\textbf{(c) Population similarity decays with decode step, not with numerical scale alone.}
Linear CKA stays at $\geq 0.995$ at the prompt for every comparison, then falls monotonically with the generated-token index, reaching 0.64 (\bft, batch size) and 0.95 (\fps, batch size) at token 20.
The decay mirrors the divergence curve of \cref{fig:acts}b, not the per-element rounding scale.

\textbf{(d) Intrinsic dimensionality is unaffected.}
The participation ratio of the two runs' activation matrices agrees to a median ratio of 1.000 (5th--95th percentile 0.97--1.03), even where CKA has fallen well below 1.
The perturbation moves points within a cloud of unchanged effective dimension rather than expanding or collapsing it.
Absolute participation ratios vary widely across models and depths (1.0--183), so we compare only within a (model, layer, position) cell.

\section{Matched probe accuracy}
\label{app:matched}

\Cref{tab:matched} lists matched-configuration test accuracy, averaged over layers and training precisions.
Prefill probes are the most accurate; accuracy declines with generation step as the probed token moves away from the input.
Pooled over all four positions, the best layers on Rotten Tomatoes are the middle, three-quarter and final blocks (83.1--84.7\%) and block 3 is the weakest (71.0\%).
Those figures average the prompt together with the three decode positions; at the prompt alone the same blocks reach 89.7--90.0\% and block 3 reaches 75.1\%, which is what \cref{tab:matched} tabulates by position.

\begin{table}[H]
\caption{Matched-configuration test accuracy (\%), mean over four layers and the training precisions available for each model.}
\label{tab:matched}
\centering
\small
\begin{tabular}{lllrrrr}
\toprule
Dataset & Model & Probe & prefill & gen 5 & gen 10 & gen 20 \\
\midrule
rotten tomatoes & Gemma-3-4B & LR & 85.5 & 81.2 & 79.5 & 77.1 \\
rotten tomatoes & Gemma-3-4B & MLP-1 & 85.4 & 80.3 & 78.5 & 76.4 \\
rotten tomatoes & Gemma-3-4B & MLP-3 & 85.6 & 80.1 & 78.6 & 77.0 \\
rotten tomatoes & Llama-3.1-8B & LR & 87.4 & 83.5 & 79.2 & 75.4 \\
rotten tomatoes & Llama-3.1-8B & MLP-1 & 86.6 & 82.7 & 77.9 & 73.6 \\
rotten tomatoes & Llama-3.1-8B & MLP-3 & 86.7 & 82.2 & 78.8 & 73.7 \\
rotten tomatoes & Qwen3-8B & LR & 86.2 & 82.7 & 80.4 & 77.4 \\
rotten tomatoes & Qwen3-8B & MLP-1 & 85.9 & 81.6 & 79.5 & 75.9 \\
rotten tomatoes & Qwen3-8B & MLP-3 & 85.7 & 80.0 & 79.2 & 76.0 \\
sms spam & Gemma-3-4B & LR & 97.7 & 96.5 & 93.4 & 91.0 \\
sms spam & Gemma-3-4B & MLP-1 & 99.3 & 98.3 & 97.6 & 96.7 \\
sms spam & Gemma-3-4B & MLP-3 & 99.4 & 98.2 & 97.4 & 96.1 \\
sms spam & Llama-3.1-8B & LR & 98.1 & 94.0 & 92.9 & 91.1 \\
sms spam & Llama-3.1-8B & MLP-1 & 99.0 & 98.3 & 97.5 & 97.1 \\
sms spam & Llama-3.1-8B & MLP-3 & 99.1 & 97.7 & 96.9 & 96.5 \\
sms spam & Qwen3-8B & LR & 97.7 & 96.5 & 93.7 & 91.6 \\
sms spam & Qwen3-8B & MLP-1 & 99.3 & 98.4 & 98.2 & 97.1 \\
sms spam & Qwen3-8B & MLP-3 & 99.3 & 98.5 & 98.0 & 96.7 \\
\bottomrule
\end{tabular}

\end{table}

\end{document}